\documentclass[10pt,twocolumn,letterpaper]{article}

\usepackage[pagenumbers]{style} % To force page 
\usepackage{graphicx}
\usepackage{amsmath}
\usepackage{amssymb}
\usepackage{booktabs}
\usepackage{multirow}

\usepackage[square,sort,comma,numbers]{natbib}
\usepackage{multibib}
\newcites{appendix}{Appendix References}

\usepackage{colortbl}
\usepackage[table]{xcolor}
\usepackage{xfp}

\usepackage{enumitem}
\setlist[itemize]{leftmargin=*}

\def\comp{\ensuremath\mathop{\scalebox{.6}{$\circ$}}}

\definecolor{G}{HTML}{39b54a} % Green color from Moco
\definecolor{R}{HTML}{ff0000} % c41e3a
\definecolor{B}{HTML}{000000}
\newcommand\resdiff[3]{\textcolor{#1}{\scriptsize{~(${#2}${#3})}}}

\definecolor{demphtcolor}{HTML}{C40000}%FF8478

\definecolor{emphtcolor}{HTML}{00B300}%84FF78

\definecolor{mygray}{gray}{0.7}

\usepackage{marvosym}
\usepackage[pagebackref,breaklinks,colorlinks]{hyperref}

\usepackage[capitalize]{cleveref}
\crefname{section}{Sec.}{Secs.}
\Crefname{section}{Section}{Sections}
\Crefname{table}{Table}{Tables}
\crefname{table}{Tab.}{Tabs.}

\def\eg{\textit{e.g.,~}}
\def\ie{\textit{i.e.,~}}

\makeatletter
\def\blfootnote{\xdef\@thefnmark{}\@footnotetext}
\makeatother
\begin{document}

\title{Uni-Perceiver v2: A Generalist Model\\for Large-Scale Vision and Vision-Language Tasks}

\author{
    Hao Li$^{1*}$, 
    Jinguo Zhu$^{2*}$, 
    Xiaohu Jiang$^{3*}$, 
    Xizhou Zhu$^{4,6}$\textsuperscript{\Letter}, 
    Hongsheng Li$^{1}$, \\
% \vspace{0.2em}
    Chun Yuan$^{3}$,
    Xiaohua Wang$^{2}$, 
    Yu Qiao$^{6}$, 
    % \vspace{0.2em}\\
    Xiaogang Wang$^1$,
    Wenhai Wang$^6$,
    Jifeng Dai$^{5,6}$
\vspace{0.1em}\\
$^{1}$CUHK-SenseTime Joint Laboratory, The Chinese University of Hong Kong \\
$^{2}$Xi'an Jiaotong University \quad
$^{3}$SIGS, Tsinghua University \quad
$^{4}$SenseTime Research \\
$^{5}$Tsinghua University \quad
$^{6}$Shanghai Artificial Intelligence Laboratory \\ 
\texttt{\small haoli@link.cuhk.edu.hk}, 
\texttt{\small lechatelia@stu.xjtu.edu.cn}, \\
\texttt{\small jiangxh21@mails.tsinghua.edu.cn},  
\texttt{\small zhuwalter@sensetime.com} \\
\texttt{\small daijifeng@tsinghua.edu.cn}, 
\texttt{\small \{hsli,\,xgwang\}@ee.cuhk.edu.hk}, \\
\texttt{\small yuanc@sz.tsinghua.edu.cn},
\texttt{\small xhw@mail.xjtu.edu.cn},
\texttt{\small \{qiaoyu,\,wangwenhai\}@pjlab.org.cn}
}
\maketitle

\thispagestyle{empty}

\blfootnote{\noindent $^{*}$Equal contribution. This work is done when Hao Li, Jinguo, and Xiaohu are interns at Shanghai Artificial Intelligence Laboratory. 
Code  shall be released at \url{https://github.com/fundamentalvision/Uni-Perceiver}.
\textsuperscript{\Letter}Corresponding author.}

%%%%%%%%% ABSTRACT
\begin{abstract}

Despite the remarkable success of foundation models, their task-specific fine-tuning paradigm makes them inconsistent with the goal of general perception modeling. The key to eliminating this inconsistency is to use generalist models for general task modeling. However, existing attempts at generalist models are inadequate in both versatility and performance. In this paper, we propose Uni-Perceiver v2, which is the first generalist model capable of handling major large-scale vision and vision-language tasks with competitive performance. Specifically, images are encoded as general region proposals, while texts are encoded via a Transformer-based language model. The encoded representations are transformed by a task-agnostic decoder. Different tasks are formulated as a unified maximum likelihood estimation problem. We further propose an improved optimizer to ensure stable multi-task learning with an unmixed sampling strategy, which is helpful for tasks requiring large batch-size training.  After being jointly trained on various tasks, Uni-Perceiver v2 is capable of directly handling downstream tasks without any task-specific adaptation. Results show that Uni-Perceiver v2 outperforms all existing generalist models in both versatility and performance. Meanwhile, compared with the commonly-recognized strong baselines that require tasks-specific fine-tuning, Uni-Perceiver v2 achieves competitive performance on a broad range of vision and vision-language tasks.

\end{abstract}

%%%%%%%%% BODY TEXT
% \input{figures/overview.tex}

\section{Introduction}
\label{sec:introduction}

\begin{figure*}[t]
\centering
\includegraphics[width=0.99\textwidth]{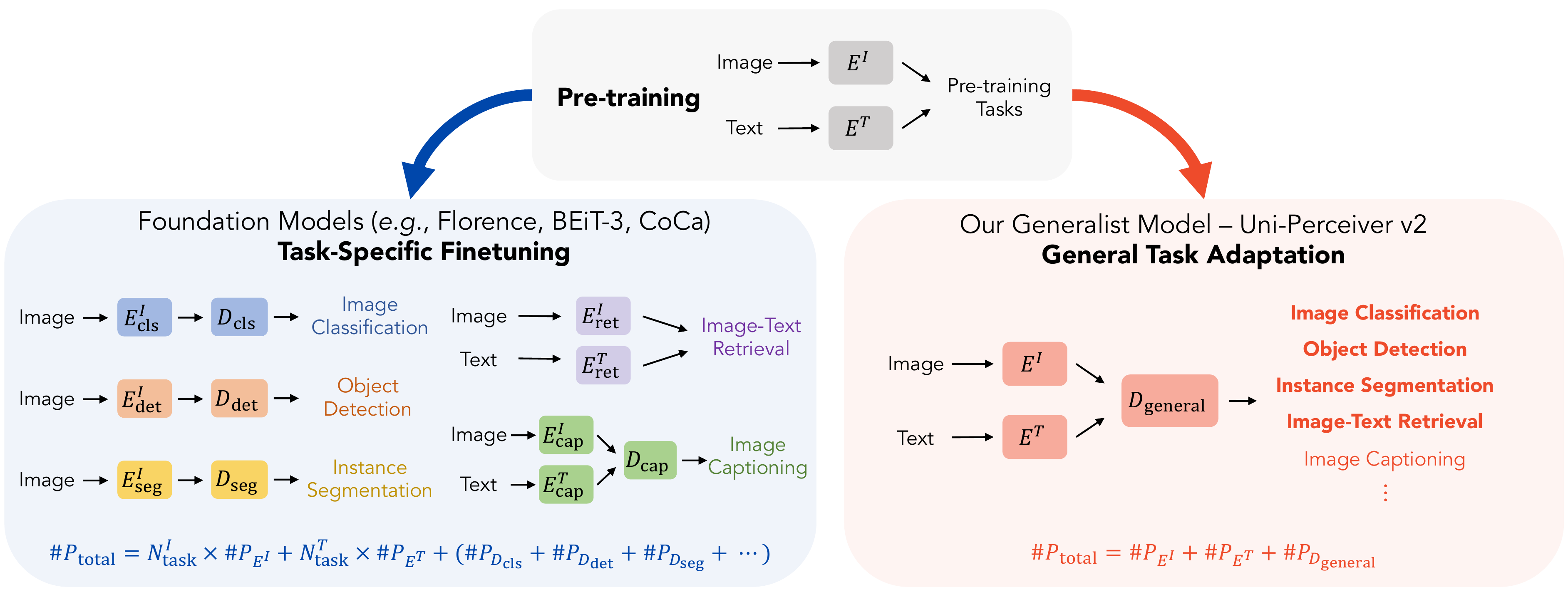}
%\vspace{-0.6em}
\caption{Comparison of foundation models and Uni-Perceiver v2. $E^I$ and $E^T$ denote the image encoder and text encoder, respectively. In existing foundation models, task-specific decoders $D_\text{cls}$, $D_\text{det}, \dots$ are employed to tune $E^I$ and $E^T$ in different task-specific finetuning. The total number of parameters $\#P_\text{total}$ in adaptation grow with the number of visual/linguistic tasks, denoted as $N^I_\text{task}$ and $N^T_\text{task}$, respectively. By contrast, our Uni-Perceiver v2 shares all parameters across various downstream tasks with a general decoder $D_\text{general}$, where no task-specific fine-tuning is incorporated. Better than previous generalist models, our method can also effectively handle pillar tasks such as image classification, object detection, instance segmentation, and image-text retrieval.}
\label{fig:paradigm}
\vspace{-1.em}
\end{figure*}

Learning a \emph{general perception model} that can handle various modalities and tasks is widely regarded as an important step towards artificial general intelligence. Due to its difficulty, many works (\eg Florence~\cite{yuan2021florence}, CoCa~\cite{yu2022coca}, BEiT-3~\cite{wang2022image}), also known as \emph{foundation models}~\cite{bommasani2021opportunities}, instead focus on a fallback solution of learning a general representation encoder that can be adapted (\eg fine-tuned) to various downstream tasks. By performing large-scale pre-training on massive multi-modal task-agnostic data, these works have demonstrated the superiority by pushing the state-of-the-art results on a broad range of tasks including single-modal tasks (\eg image classification and object detection) and also cross-modal tasks (\eg image captioning and image retrieval).

Despite the success, there is still a considerable gap between foundation models and the goal of general perception modeling. While foundation models only focus on general representation learning, task modeling is neglected. Traditional task-specific fine-tuning paradigm is still utilized (see Fig.~\ref{fig:paradigm}). This significantly increases the marginal cost of adapting pre-trained models to various downstream tasks, making it difficult to meet the rapidly growing demands of diverse downstream tasks and scenarios. Such a task-specific fine-tuning paradigm of foundation models is inconsistent with the goal of general perception modeling.

Instead of performing task-specific fine-tuning, generalist models process different tasks with shared architecture and parameters, which is aligned with the goal of general perception modeling. It not only reduces the cost of handling diverse tasks but also enables task collaboration. Most existing attempts on generalist models are sequence-to-sequence (seq2seq) models~\cite{pix2seqv2,ofa,unifiedio,alayrac2022flamingo,gpv1,gpv2,gato,unitab}. However, these attempts are inadequate in both versatility and performance: (1) some pillar vision and vision-language tasks as listed in Tab.~\ref{tab:pillar_tasks} cannot be handled, \eg image-text retrieval, object detection, and instance segmentation; (2) the accuracy and inference speed still lag significantly behind state-of-the-art task-specific methods. Another line of research named Uni-Perceivers~\cite{zhu2022uni,zhu2022uni_p} builds generalist models supporting both generation and non-generation tasks. Nevertheless, they still cannot handle many vital tasks such as detection and segmentation.

To develop generalist models with better versatility and performance, our core idea is to encode images as general region proposals consisting of the semantic, bounding box and segmentation mask representations. Compared with previous methods where images are represented as non-overlapping patches, this design makes our localization modeling more expressive and flexible. This explicit utilization of localization clues not only greatly reduces the difficulty of handling localization tasks such as image detection and segmentation, but also provides richer features for non-localization tasks, thus enabling more general task modeling and better performance.

In this paper, we propose Uni-Perceiver v2 as a generalist model capable of handling major large-scale vision and vision-language tasks as listed in Tab.~\ref{tab:pillar_tasks}. Specifically, images are encoded as a concatenation of global and regional representations via a region proposal network, while texts are encoded via a Transformer-based language model. Both the image and text encoders can benefit from off-the-shelf pre-trained models, which reduces the demand for training data and resources and ensures performance. The encoded representations are transformed by a shared modality-agnostic Transformer~\cite{vaswani2017attention} network to obtain the decoded representations. Following Uni-Perceivers~\cite{zhu2022uni_p,zhu2022uni}, different tasks are formulated as a unified maximum likelihood estimation problem and are jointly learned to enable general task adaptation. We further propose an improved optimizer named MT-AdamW to ensure stable multi-task learning with an unmixed sampling strategy which only samples one task for all GPUs per iteration. This is very helpful for tasks requiring large batch size training.

Uni-Perceiver v2 is the first generalist model achieving competitive results on major large-scale vision and vision-language tasks including object detection, instance segmentation, image classification, image captioning, and image-text retrieval, except for image generation that has not been verified due to limited computational resources. After being jointly trained on various tasks, it can directly handle a broad range of tasks without any task-specific adaption, achieving state-of-the-art performance among existing generalist models.
Our contributions are summarized as:
\begin{itemize}
\vspace{-0.4em}
\item We propose Uni-Perceiver v2, which is the first generalist model capable of handling both localization and non-localization tasks with competitive performance. The general region proposal encoding of images brings more flexible and expressive localization modeling.
\vspace{-0.4em}
\item To improve the effectiveness of multi-task learning, we adopt an unmixed sampling strategy to enable large batch-size training and develop an improved optimizer named MT-AdamW to mitigate the instability in gradients. 
\vspace{-0.4em}
\item Uni-Perceiver v2 outperforms all existing generalist models in both versatility and performance. Without any task-specific adaption, Uni-Perceiver v2 achieves competitive performance on a broad range of downstream tasks compared with commonly-recognized strong baselines that require task-specific fine-tuning, demonstrating its strong ability of general task modeling.

\end{itemize}

\begin{table}
\centering
\small
\resizebox{0.8\linewidth}{!}{
\begin{tabular}{c|c}
    \toprule
  Categories  & Specific Tasks  \\
\midrule
  \multirow{2}{*}{Retrieval} &\textbf{\multirow{2}{*}{\underline{Image-text retrieval}}}   \\
   &    \\
 \midrule
   \multirow{3}{*}{Classification} & \textbf{\underline{Image classification}}  \\
   &   \textcolor{gray}{Region categorization} \\
   &   \textcolor{gray}{Situation recognition} \\
\midrule
\multirow{8}{*}{Localization} & \textbf{\underline{Object detection}} \\
& \textcolor{gray}{Key point detection}  \\
& \textcolor{gray}{Pose estimation} \\
& \textcolor{gray}{Referring expression grounding}  \\
& \textcolor{gray}{Human object interaction}  \\
& \textcolor{gray}{Relation detection}  \\
& \textcolor{gray}{Optical character recognition}  \\
& \textcolor{gray}{Object localization}  \\
\midrule
\multirow{3}{*}{Mask Predication} & \textbf{\underline{Instance segmentation}}  \\
& \textcolor{gray}{Semantic segmentation}   \\
& \textcolor{gray}{Panoptic segmentation} \\
\midrule
\multirow{8}{*}{Image Generation} & \textbf{Image synthesis} \\
& \textcolor{gray}{Image inpainting}  \\
& \textcolor{gray}{Segment-based image generation}  \\
& \textcolor{gray}{Style transferring}  \\
& \textcolor{gray}{Depth estimation}  \\
& \textcolor{gray}{Surface normal estimation}   \\
& \textcolor{gray}{Image infilling}  \\
& \textcolor{gray}{Image super resolution}  \\
\midrule
\multirow{6}{*}{Image to Text} & \textbf{\underline{Image captioning}}  \\
& \textcolor{gray}{Visual question answering}   \\
& \textcolor{gray}{Region captioning}   \\
& \textcolor{gray}{Grounded VQA}   \\
& \textcolor{gray}{Grounded captioning}   \\
& \textcolor{gray}{Visual commonsense reasoning}   \\

   \bottomrule   
    \end{tabular}
    }
\caption{Categories of mainstream vision and vision-language tasks. Pillar tasks of different downstream task categories are in \textbf{bold}. These pillar tasks are the most representative tasks in each category, where other tasks can be derived from them. Uni-Perceiver v2 is able to effectively handle the \underline{underlined} pillar tasks, except for image synthesis that has not been verified due to limited computational resources.}
\vspace{-1em}
\label{tab:pillar_tasks}
\end{table}

\section{Related Work}
\label{sec:related_work}

\noindent\textbf{Foundation Vision Models~} are ``designed to be adapted (\eg fine-tuned) to various downstream tasks by \emph{pre-training} on broad data at scale''~\cite{bommasani2021opportunities}. Such large-scale pre-trained vision models have shown effectiveness in enriching data encoding capacity, alleviating data hunger, and improving the performance of downstream tasks.

Image classification on ImageNet-1k~\cite{deng2009imagenet} has been the mainstream pre-training paradigm for a long period. 
However, as the model size grows, larger annotated datasets are required to avoid over-fitting in pre-training, such as ImageNet-21k~\cite{deng2009imagenet}, Instagram-1B~\cite{mahajan2018exploring}, JFT-300M~\cite{sun2017revisiting} and JFT-3B~\cite{zhai2022scaling}.
Inspired by the success of linguistic pre-training on massive web-crawled text, CLIP~\cite{clip} and ALIGN~\cite{align} have begun to focus on multi-modal contrastive pre-training on web-scale noisy image-text pairs to learn aligned image and text representations.  SimVLM~\cite{wang2021simvlm} employs the multi-modal sequence generation task for pre-training. 
FLAVA~\cite{singh2021flava} combines contrastive and generative pre-training to handle both unimodal and multimodal tasks.
UniCL~\cite{yang2022unified} and CoCa~\cite{yu2022coca} jointly use human-annotated and web-crawled data. 
Florence~\cite{yuan2021florence} and INTERN~\cite{shao2021intern} increase the scale and diversity of pre-training data to enhance the representation capability.
OmniVL~\cite{wang2022omnivl} proposes to incorporate both image-language and video-language tasks in its pre-training.
BEiT-3~\cite{wang2022image} unifies pre-training objectives for different modalities as a single masked data modeling task, achieving state-of-the-art results on a wide range of downstream tasks.

These works on foundation models only focus on general representation learning, while neglecting task modeling. When adapting them to downstream tasks, the traditional task-specific fine-tuning paradigm is still utilized, which is inconsistent with the goal of general perception modeling. Meanwhile, with the rapidly growing demands of diverse tasks and scenarios, the task-specific fine-tuning paradigm would result in a prohibitive marginal cost for data collection, data annotation, model training, and model storage.

\vspace{0.5em}\noindent\textbf{Generalist models~} handle various tasks with shared architecture and parameters, which have been long pursued by the machine learning community. 
Recently, inspire by the success of sequence-to-sequence (seq2seq) models in NLP field~\cite{radford2018improving},  OFA~\cite{ofa}, Flamingo~\cite{alayrac2022flamingo}, and GIT~\cite{wang2022git} propose to model various tasks as a sequence generation task.
Unified-IO~\cite{unifiedio}, Pix2Seq v2~\cite{pix2seqv2}, and UniTab~\cite{unitab} further develop this method to support more tasks by introducing discrete coordinate tokens, thus location information can be encoded or decoded by the unified models.
Beyond that, Gato~\cite{gato} succeeds in unifying reinforcement learning tasks into the seq2seq framework.
GPV~\cite{gpv1} also builds a general-purpose vision system by adding a seq2seq module on a DETR~\cite{carion2020end}-based visual encoder.

However, these methods with seq2seq formulation are still inadequate in both versatility and performance:
(1) They cannot handle some core vision tasks, \eg image-text retrieval, object detection, and instance segmentation. Although Pix2Seq v2~\cite{pix2seqv2} includes detection and instance segmentation tasks, its performance and inference speed still lag significantly behind state-of-the-art task-specific methods~\cite{zhang2022dino,li2022mask};
(2) The non-parallel auto-regressive decoding leads to slow inference speed. For example, image classification requires calculating and comparing the cumulative probabilities of all category names conditioned on the given image;
(3) They also suffer from the task-interference issue in multi-task learning, resulting in performance degradation compared with task-specific models.

Alternatively, Uni-Perceivers~\cite{zhu2022uni_p,zhu2022uni} formulate different tasks as finding the maximum likelihood target for each input through the representation similarity regardless of their modality,  making it possible to support both generation and non-generation tasks. Nevertheless, they still cannot handle image detection and segmentation tasks.

\section{Revisiting Uni-Perceivers}

\vspace{0.5em}\noindent\textbf{Unified Modeling of Perception Tasks.~}
Uni-Perceiver~\cite{zhu2022uni_p} proposes to reformulate different tasks as a unified maximum likelihood estimation problem. 
Specifically, each task is defined with a set of inputs and a set of candidate targets from arbitrary combinations of modalities. The inputs and targets are first encoded with a modality-specific tokenizer with linear projection. Then the encoded representations are transformed by modality-agnostic decoder with shared parameters for different tasks. Given an input, the unified task objective is defined as finding the target with the maximum likelihood with the input. 

\vspace{0.5em}\noindent\textbf{Mitigating Task Interference.~}
Multi-task learning with fully shared parameters could introduce interference between different tasks.
Uni-Perceiver-MoE~\cite{zhu2022uni} proposes Conditional MoEs to address the task-interference issue. Specifically, for each input token, a routing decision is calculated depending on specific routing strategy, which sparsely activates a small portion of experts to process this token. The corresponding output of an input token is the linearly weighted combination of those selected experts by the routing decision. Conditional MoEs mitigate the interference issue by allowing conflicting modalities and tasks using separate parameters without introducing any task-specific modules.

\vspace{0.5em}\noindent\textbf{Limitations.~}
Although Uni-Perceivers aim to process different tasks with a unified architecture, it fails to handle detection and segmentation tasks due to the lack of localization information in its encoded features. Meanwhile, Uni-Perceivers do not integrate off-the-shelf encoder models, making it unable to benefit from existing large-scale pre-trained encoders. This potentially increases its demand for pre-training data and resources, limiting its performance.

\section{Method}
\label{sec:method}

\subsection{Encoding Images as General Region Proposals}
\label{sec:method:imageenc}

Most existing generalist models~\cite{zhu2022uni_p,zhu2022uni} represent images as non-overlapping patches with fixed sizes. This design is rather coarse and limited in modeling objects of varying sizes and shapes in images, making it difficult to handle localization tasks such as detection and segmentation.

In order to enable more expressive and flexible localization modeling, we propose to encode the input image as a sequence of general region proposals. Specifically, given an input image $x \in \mathbb{R}^{H \times W}$ with height $H$ and width $W$, a network $f_{\text{image}}(\cdot)$ is employed to encode the image as the concatenation of global and regional representations as
\begin{equation}
    f_\text{image}(x) = \text{Concat}\left(\{q_i^\text{global}\}_{i=1}^M ~,~ \{q_j^\text{proposal}\}_{j=1}^N\right),
\end{equation}
where $q_i^{\text{global}} \in \mathbb{R}^{d}$ are the global representations of the whole image, and $q_j^{\text{proposal}} \in \mathbb{R}^{d}$ are the regional representations of candidate object proposals in the image.

Following the common practice in localization tasks, an image backbone network (\eg ResNet~\cite{he2016deep}) is firstly employed to extract the multi-scale feature maps $\{\mathcal{F}_l\}_{l=1}^L$, where $L$ is the number of feature scales (\eg $L=4$).

\vspace{0.5em}\noindent\textbf{Regional Representations.~} A Transformer~\cite{vaswani2017attention}-based region proposal network is applied on top of the multi-scale feature maps $\{\mathcal{F}_l\}_{l=1}^L$ to extract a set of $O$ candidate object proposals $\{ q_j^{\text{sem}}, q_j^{\text{box}}, q_j^{\text{mask}} \}_{j=1}^O$, where $q_j^{\text{sem}} \in \mathbb{R}^d$, $q_j^{\text{box}} \in \mathbb{R}^4$, and $q_j^{\text{mask}} \in \mathbb{R}^{H \times W} $ are the semantic, bounding box, and segmentation mask representations of the $j$-th proposal, respectively. The region proposal network is similar to MaskDINO~\cite{li2022mask}, but only considers foreground-background binary classification. See Appendix for detailed implementation. These three representations are then fused as the regional representation as
\begin{equation}
    q_j^{\text{proposal}} = q_j^{\text{sem}} + \mathcal{B}(q_j^{\text{box}}) + \mathcal{M}(q_j^{\text{mask}}),
\end{equation}
where $\mathcal{B}$ denotes the positional encoding of box coordinates.
$\mathcal{M}$ uses adaptive average pooling to scale the mask predictions to the size of $28\times28$. Both $\mathcal{B}$ and $\mathcal{M}$ are followed by linear projections to match the feature dimension.

\vspace{0.5em}\noindent\textbf{Global Representations.} The global representations are extracted from the last-scale feature map $\mathcal{F}_L \in \mathbb{R}^{h \times w}$ with height $h$ and width $w$. $M'$ instances of parameterized Attention Pooling~\cite{clip} are employed to extract global features. The pooled features are concatenated with the flattened feature map to obtain the global representations as
\begin{equation}
    q^{\text{global}} = \text{Concat}\Big(\big\{\operatorname{AttnPool}_i(\mathcal{F}_L)\big\}_{i=1}^{M'}~,~ \operatorname{Flatten}(\mathcal{F}_L)\Big).
\end{equation}

\subsection{Encoding Text with Language Models}
\label{sec:method:textenc}
A Transformer~\cite{vaswani2017attention}-based language model is used to encode textual data, such as category names in classification tasks, image descriptions in image-text retrieval tasks, and the vocabulary in image captioning tasks.
Specifically, a BPE tokenizer~\cite{sennrich2015neural} tokenizes the input text $x$ into a sequence of word embeddings, and a Transformer encoder is employed to extracts the text feature sequence as
\begin{equation}
    f_{\text{text}}(x) = \text{Concat}(q_1^\text{text}, q_2^\text{text}, \cdots, q_L^\text{text}) 
\end{equation}
where $q_i^\text{text}\in  \mathbb{R}^d$ is the encoded feature of the $i$-th word, and $L$ is the sequence length.
In our implementation, we use a pre-trained RoBERTa$_\text{~BASE}$~\cite{liu2019roberta} as the text encoder, which is jointly tuned with the whole network.

\subsection{General Task Adaptation}
We follow Uni-Perceivers~\cite{zhu2022uni_p,zhu2022uni} to formulate different tasks as a unified maximum likelihood estimation problem. Given an input $x \in \mathcal{X}$ and the candidate target set $\mathcal{Y}$, the task objective is defined as finding the target $\hat{y} \in \mathcal{Y}$ with the maximum likelihood as
\begin{equation}
    \hat{y} = \arg\max_{y \in \mathcal{Y}} P(x, y),
    \label{eq1}
\end{equation}
where the likelihood $P(x, y)$ is estimated from the cosine similarity between the representations of $x$ and $y$ as
\begin{equation}
    P(x, y) \propto \exp\bigg(\cos\Big( g\comp f(x) ~,~ g\comp f(y) \Big) / \tau \bigg),
\end{equation}
where $f(\cdot)$ is the modality-specific encoders $f_\text{image}$ and $f_\text{text}$ introduced in Sec.~\ref{sec:method:imageenc} and \ref{sec:method:textenc}, respectively. $g(\cdot)$ is a modality-agnostic Transformer~\cite{vaswani2017attention} network shared for different tasks, and $\tau > 0$ is a learnable temperature parameter. 

Depending on task requirements, the modality-specific encoded representation for inputs $x$ can be an image feature sequence $f_{\text{image}}(x)$, a text feature sequence $f_{\text{text}}(x)$, or their concatenation, with an additional \verb|<SPE>| token inserted at the beginning. The encoded representation for targets $y$ is constructed in the same way.

To obtain general task modeling capability, Uni-Perceiver v2 conducts multi-task learning on various uni-modal and multi-modal tasks. Denoting a set of $K$ tasks as $\{\mathcal{X}_k, \mathcal{Y}_k\}_{k=1}^{K}$, where $\mathcal{X}_k$ and $\mathcal{Y}_k$ are the input set and target set of the $k$-th task, respectively. The training loss is 
\begin{equation}
\small
L = \sum_{k=1}^{K} s_k \mathop{\mathbb{E}}_{\{x, y\} \in \{\mathcal{X}_k, \mathcal{Y}_k\}} \bigg[- w_k\log \frac{P(x, y)}{\sum_{z \in \mathcal{Y}_k}{P(x, z)}}\bigg],
\label{old loss}
\end{equation}
where $s_k$ and $w_k$ denote the sampling ratio and loss weight of the $k$-th task, respectively. The sampling ratio are normalized as $\sum_k s_k = 1$. We refer to Sec.~\ref{sec:method:samplestrategy} for detailed discussions of the sampling strategy. To mitigate the task interference in multi-task training, we follow Uni-Perceiver-MoE~\cite{zhu2022uni} to employ the Conditional MoEs with attribute-level routing strategy for effective multi-task training.

\vspace{0.5em}\noindent \textbf{Tasks with Localization.}
Uni-Perceiver v2 can perform localization tasks such as object detection and instance segmentation by decoding the regional representations.
Specifically, for each region proposal $q_j^{\text{proposal}}$, its outputted feature from the unified decoder $g(\cdot)$ will be compared with class embeddings to obtain the class prediction as in Eq.~\eqref{eq1}. The corresponding bounding box $q_j^{\text{box}}$ and segmentation mask $q_j^{\text{mask}}$ will serve as the localization predictions.

\vspace{0.5em}\noindent \textbf{Tasks without Localization.}
Uni-Perceiver v2 can also handle tasks that do need localization predictions, \eg
image classification, image captioning, image-text retrieval. 
It follows a similar formulation of Uni-Perceiver for these tasks with two major differences:
(1) More expressive and flexible localization clues for images, better facilitating these tasks;
(2) Both the image and text encoders can leverage off-the-shelf modality-specific pre-trained models, leading to better performance.

\subsection{Sampling Strategy and Improved Optimization }
\label{sec:method:samplestrategy}

Optimizing generalist models follows the paradigm of multi-task learning, which performs joint training on data from different tasks.
Current methods usually mix all tasks in one training iteration~\cite{zhu2022uni_p,ofa,unifiedio}.
Such \textit{mixed sampling strategy} limits the batch-size of each task, which can be detrimental for tasks that benefit from large batch-size training (\eg image-text retrieval).

A straightforward solution is to sample only one task per iteration, which we refer as \textit{unmixed sampling strategy}. It can achieve the largest training batch-size.
However, when different iterations sample different tasks, the gradients would vary greatly due to the differences in data and tasks, which may bring potential instability to multi-task learning and performance deterioration.

To mitigate the instability issue of unmixed sampling strategy, we propose an improved optimizer for multi-task training, named as \textbf{MT-AdamW}. The core idea is to balance the gradient of each task, by normalizing the gradient of each iteration and compensating it according to the task sampling ratio. 

Suppose the $k$-th task is sampled at timestep $t$, the vanilla AdamW~\cite{loshchilov2017decoupled} is modified to MT-AdamW by updating the parameters $\theta$ as follows:
\begin{equation*}
\small
\left\{\begin{array}{l}
\mathbf{g}_{t} \leftarrow  {\nabla L_{t,k}\left(\mathbf{\theta}_{t-1}\right)}  \\
\mathbf{m}_t{=}{(}1{-}\beta_1{)} \mathbf{m}_{t{-}{1}}{+}\beta_1\mathbf{g}_{t} \\
\mathbf{n}_t{=}{(}1{-}\beta_2{)} \mathbf{n}_{t{-}{1}}{+}\beta_2 \mathbf{g}_{t}^2 \\
\theta_{t}{=}\theta_{t-1}-\alpha \frac{\mathbf{m}_t }{\sqrt{\mathbf{n}_t}+\varepsilon } 
\end{array}\right.
\Rightarrow
\left\{\begin{array}{l}
\mathbf{g}_{t} \leftarrow  {\color{red}\omega_k} \frac{\nabla L_{t,k}\left(\mathbf{\theta}_{t-1}\right)}{\color{red} \lVert \nabla L_{t,k}\left(\mathbf{\theta}_{t-1}\right) \rVert}  \\
\mathbf{m}_t{=}{(}1{-}\beta_1{)} \mathbf{m}_{t{-}{1}}{+}\frac{\beta_1}{\color{red}s_k} \mathbf{g}_{t} \\
\mathbf{n}_t{=}{(}1{-}\beta_2{)} \mathbf{n}_{t{-}{1}}{+}\frac{\beta_2}{\color{red}s_k} \mathbf{g}_{t}^2 \\
\theta_{t}{=}\theta_{t-1}-\alpha \frac{\mathbf{m}_t }{\sqrt{\mathbf{n}_t}+\varepsilon } 
\end{array}\right.
%\label{newoptim}
\end{equation*}
where $L_{t,k}$ is the loss function for the sampled $k$-th task at timestep $t$, and $\alpha$ is the learning rate. The weight decay and bias corrections are omitted for simplicity. The original task gradients are first normalized to stabilize training. The scaling factor $\omega_k$ serves as the loss weight of the sampled task. Then the trimmed gradient $\mathbf{g}_{t}$ can be used to estimate the first moment $\mathbf{m}_t$ and second moment $\mathbf{n}_t$ of gradients in a moving average way. 
To further decouple the gradient contribution and sampling ratio $s_k$ of each task, a task-specific compensation coefficient $1 / s_k$ is used to unbias the estimation $\mathbf{m}_t$ and $\mathbf{n}_t$. In practice, if all tasks are expected to contribute equally, all scaling factors could be set as $\omega_k=1$.

\section{Experiments}
\label{sec:experiments}

\subsection{Datasets}
Uni-Perceiver v2 performs multi-task training on various tasks and public-available datasets to achieve the general task modeling capability.
It uses similar datasets as in Uni-Perceiver~\cite{zhu2022uni_p}. 
Specifically, the image classification task is trained on ImageNet-1k~\cite{deng2009imagenet} dataset. 
For objection detection and instance segmentation, COCO~\cite{lin2014microsoft} is used for training.
For image captioning and image-text retrieval, we use a combination of image-text-pair datasets: SBU Captions~\cite{ordonez2011im2text}, Visual
Genome~\cite{krishna2017visual}, COCO Caption~\cite{Chen2015MicrosoftCC}, CC3M~\cite{sharma2018conceptual}, CC12M~\cite{changpinyo2021cc12m} and YFCC~\cite{yfcc}. 
We also add the language modeling task during training, which is trained on BookCorpus~\cite{zhu2015aligning} and English Wikipedia (Books\&Wiki).

During the evaluation, we evaluate generalist models on the most representative datasets for the pillar vision and vision-language tasks listed in Tab.~\ref{tab:pillar_tasks}. Specifically, ImageNet-1k~\cite{deng2009imagenet} and COCO Caption~\cite{Chen2015MicrosoftCC} are utilized to evaluate the performance of image classification and image caption, respectively. 
For image-text retrieval, COCO Caption and Flickr30k~\cite{plummer2015flickr30k} are utilized. Note that Flickr30k is not involved in training.
For objection detection and instance segmentation, COCO~\cite{lin2014microsoft} is used to evaluate their performances.
We put the licenses of all datasets in the Appendix.

\subsection{Implementation Details}

We implement three Uni-Perceiver v2  variants with  different backbones, \ie ResNet-50~\cite{he2016deep}, Swin-Base~\cite{liu2021swin}, and Swin-Large. 
ResNet-50 is pre-trained on ImageNet-1k,  and Swin-Base is pre-trained on ImageNet-21k. 
Swin-Large is firstly pre-trained on  ImageNet-21k  and then trained on the detection task with Object365~\cite{shao2019objects365}.
The number of feature scales $L$ is set to 4 for all models.
A Transformer~\cite{vaswani2017attention}-based region proposal network is used to generate general region proposals, whose architecture and settings mainly follow Mask DINO~\cite{li2022mask}. 
However, we replace all multi-category classifiers with binary classifiers. 
In addition,  the number of global attention pooling to extract global features is set to $M' = 10$.
We choose the pre-trained RoBERTa$_\text{~BASE}$ \cite{liu2019roberta} as the text encoder, which is jointly tuned with the whole network.
The unified decoder is also a Transformer-based network, whose  parameters are initialized randomly and optimized from scratch. 
Its architecture follows the  setting of the BERT$_\text{~BASE}$~\cite{devlin2018bert} model, but it only consists of 6 Transformer layers.
To mitigate the task interference issue in multi-task learning, we also employ the attribute-level Conditional MoE~\cite{zhu2022uni} in all FFN layers of the unified decoder.
Please refer to the Appendix for more details.

Unless specifically stated, we adopt the unmixed sampling strategy, which only samples one task for all GPUs per iteration. The MT-AdamW optimizer with a base learning rate of 0.0001 and a weight decay of 0.0001 is utilized.
The learning rate of  modality-specific encoders is multiplied by 0.1 since they have already been pre-trained.
Uni-Perceiver v2 with Swin-Base and Swin-Large backbone is trained for 200,000 iterations on 32 and 64 NVIDIA A100 GPUs, respectively.
The learning rate drops to $0.1\times$ at the 160,000 iterations. 
For models with ResNet-50, we only train them on 16 NVIDIA A100 GPUs for 150,000 iterations. For other training settings, please also refer to the Appendix.

\subsection{Ablation Studies}
In the following, we evaluate the key components of Uni-Perceiver v2 with ResNet-50 backbone by evaluating its performance on four tasks, \ie image detection on COCO, image classification on ImageNet-1k, image-text retrieval on COCO caption, and image captioning on COCO caption.
The instance segmentation and language modeling tasks are not included to save training costs, and the YFCC dataset is also excluded from the training.
Note that, the performance on these datasets are reported without any task-specific fine-tuning.
If not stated, COCO detection pre-trained ResNet-50 is used for ablation studies to accelerate the convergence of multi-task training.

\vspace{0.5em}\noindent \textbf{Effectiveness of Global and Regional Image Representations.}
Uni-Perceiver v2 encodes images as the concatenation of global and regional representations. To evaluate their effectiveness on different tasks, we conduct experiments that employ different representations, \ie only using global representations, only using regional representation only, and using both.
Results in Tab.~\ref{tab:type of features} show that: (1) regional representation is crucial for both captioning and retrieval tasks. We speculate that this is because regional proposals can provide localization clues, which is helpful to process both tasks. (2) Compared with regional-only representations, global representations deliver better results on the image classification task, which indicates global representations are important for image-level tasks. (3) Combining global and regional representation allows the two representations to complement each other, and thus achieve the best overall results on all tasks.
Therefore, in our subsequent experiments, combining global and regional representations is taken as the default setting.
 
\setlength{\tabcolsep}{3pt}
\renewcommand{\arraystretch}{1.1}
\begin{table}[t]
  \centering
  \resizebox{0.9\linewidth}{!}{
  \begin{tabular}{c|c|c|cc|c}
    \toprule
    Representation & COCO & ImageNet-1k & \multicolumn{2}{c|}{COCO} & COCO \\
    Types & Detection & Classification & \multicolumn{2}{c|}{Retrieval}  & Caption 
 \\
    \midrule
    Global   & -  & 76.8 & 46.3& 34.6 &  28.8  \\
    Regional  & 48.2  & 75.9  & \textbf{52.3} & \textbf{39.2} & \textbf{31.2}   \\
        Global + Regional & \textbf{49.9}  & \textbf{76.9}  & 51.3 & 38.8 & 30.6  \\
    \bottomrule
  \end{tabular}
  }
  \caption{Ablation of different representation types for general region proposals. Results are reported on object detection (mAP), image classification (Acc), image-text retrieval (I2T R@1 and T2I R@1), and image caption (BLEU-4).}
  \label{tab:type of features}
  \vspace{-1em}
\end{table}

\vspace{0.5em}\noindent \textbf{Task Collaboration and Interference.}
To analyze the collaboration and interference between different tasks, we conduct experiments by removing each task independently from the joint-training tasks in Tab.~\ref{tab:task_collab}. If the removal of one task can improve (or degrade) the performance of another task, it can reflect that the former task is detrimental (or beneficial) to the latter one during joint training. For a fair comparison, the Conditional MoEs are not employed except for the last experiment. Results show that without MoEs, other tasks have negative impacts on the training of image-text retrieval. However, the image-text retrieval task could promote the performance of image captioning. The image classification task is also very helpful to image captioning, yet the reverse has no obvious effect.
It should be noted that all models employ an image encoder pre-trained on COCO detection, thereby all these tasks can benefit from the pre-trained region proposal network.
The results indicate that task interference indeed exists in the multi-task training of generalist models and is more common than task collaboration, suggesting the importance of addressing the task interference issue. By employing Conditional MoEs, the task interference is largely mitigated, resulting in improved results on all tasks.

\setlength{\tabcolsep}{3pt}
\renewcommand{\arraystretch}{1.1}
\begin{table}
  \centering
  \resizebox{\linewidth}{!}{
  \begin{tabular}{l|c|c|cc|c}
    \toprule
    \multirow{2}{*}{Tasks} & COCO & ImageNet-1k & \multicolumn{2}{c|}{COCO} & COCO \\
    & Detection & Classification & \multicolumn{2}{c|}{Retrieval}  & Caption 
 \\
    \midrule
        \textcolor{gray}{Single Task}                 & \textcolor{gray}{50.1}   & \textcolor{gray}{76.1} & \textcolor{gray}{50.0} & \textcolor{gray}{37.6} & \textcolor{gray}{30.2} \\
    \midrule
    All Tasks                           & 49.8                      & 76.3                      & 46.0                      & 34.7                      & 28.9    \\
    {\scriptsize w/o} Detection         & -                         & 76.6\resdiff{B}{+}{0.3}   & 47.0\resdiff{G}{+}{1.0}   & 34.6\resdiff{B}{-}{0.1}   & 30.4\resdiff{G}{+}{0.5}    \\
    {\scriptsize w/o} Classification  & 50.1\resdiff{B}{+}{0.3}   & -                         & 51.6\resdiff{G}{+}{5.6}   & 38.6\resdiff{G}{+}{3.9}   & 25.9\resdiff{R}{-}{3.0}   \\
    {\scriptsize w/o} Retrieval       & 49.5\resdiff{B}{-}{0.3}   & 76.3\resdiff{B}{+}{0.0}   & -                         & -                         & 27.4\resdiff{R}{-}{1.5}        \\
    {\scriptsize w/o} Captioning         & 49.7\resdiff{B}{-}{0.1}   & 76.3\resdiff{B}{+}{0.0}   & 51.2\resdiff{G}{+}{5.2}   & 38.3\resdiff{G}{+}{3.6}  & -   \\
    \midrule
    All Tasks {\scriptsize w/} MoE              & 49.9\resdiff{B}{+}{0.1}   & 76.9\resdiff{G}{+}{0.6}   & 51.3\resdiff{G}{+}{5.3}   & 38.8\resdiff{G}{+}{4.1}   & 30.6\resdiff{G}{+}{0.7} \\
    \bottomrule
  \end{tabular}
  }
  \caption{Ablation of collaboration and interference between tasks. All experiments except for the last line do not employ Conditional MoEs. In the brackets are the gaps to the ``All Tasks'' counterpart. In \textcolor{G}{green} and \textcolor{R}{red} are the gaps of at least $\pm$0.5 point.}
  \label{tab:task_collab}
  \vspace{-0.5em}
\end{table}

\setlength{\tabcolsep}{3pt}
\renewcommand{\arraystretch}{1.1}
\begin{table}
  \centering
  \resizebox{\linewidth}{!}{
  \begin{tabular}{ccc| c|c|cc|c}
    \toprule
 Task &  Gather   & MT-AdamW   & COCO & ImageNet-1k & \multicolumn{2}{c|}{COCO} & COCO \\
  Sampling   & Feature & Optimizer &  Detection & Classification & \multicolumn{2}{c|}{Retrieval}  & Caption \\
    \midrule
              mixed  &   &    &  49.6 & 76.7  & 40.1 & 31.9   & 27.6 \\
              unmixed  &  &      & 49.2  & 76.6& 39.8 & 30.9 & 27.5 \\
              % $\surd$ &  $\surd$  &    & & & & &  \\
               unmixed    & $\checkmark$        & & 49.3 & 76.8 & 50.4 & 37.3 & 27.6 \\
                \textbf{unmixed}    & \textbf{$\checkmark$} & \textbf{$\checkmark$}    & \textbf{49.9} & \textbf{76.9} & \textbf{51.3} & \textbf{38.8} & \textbf{30.6} \\
   
    \bottomrule
  \end{tabular}
  }
  \caption{Ablation of sampling strategies and improved optimizer. ``mixed'' means mixing different tasks' data in one iteration, while ``unmixed'' denotes that only one task's data is sampled in one iteration.
   ``Gather Feature'' means that negative samples for retrieval tasks are collected synchronously across GPUs.}
  \label{tab:sample strategy}
  \vspace{-1em}
\end{table}

\vspace{0.5em}
\noindent \textbf{Sampling Strategy and Improved Optimization.} 
We evaluate the effectiveness of the unmixed sampling strategy (\ie sampleing one task for each iteration) and the proposed MT-AdamW optimizer in Tab.~\ref{tab:sample strategy}. 
From the results, we observe that the vanilla unmixed sampling strategy that computing the contrastive loss with samples on each GPU have slightly adverse effect on the learning of all tasks when compared with the mixed sampling strategy. With the batch size increased by gathering features across all GPUs, the performance of retrieval tasks can be largely improved. Further introducing the MT-AdamW optimizer leads to more stable multi-task training and consistently improved performance across all tasks.

\setlength{\tabcolsep}{3pt}
\renewcommand{\arraystretch}{1.1}
\begin{table}
  \centering
  \resizebox{\linewidth}{!}{
  \begin{tabular}{c|c|c|c|cc|c}
    \toprule
    Pretrained & Pretrained & COCO & ImageNet-1k & \multicolumn{2}{c|}{COCO} & COCO \\
    Method & Data &  Detection & Classification & \multicolumn{2}{c|}{Retrieval}  & Caption 
 \\
    \midrule
    Supervised & IN-1k & 45.7 & 76.8 & 51.2 & 38.9 & 27.3 \\
    Supervised & IN-21k & 48.3 & \textbf{80.1}   & 55.1 & 41.2 & 30.2 \\
    Supervised & IN-1k \& COCO  & \textbf{49.9} & 76.9 & 51.3 & 38.8 & 30.6     \\
    MoCo v2 & IN-1k & 48.3 & 75.0  & 54.8 & 40.5 & 29.6  \\
    CLIP &  CLIP data & 47.2 & 73.8 & \textbf{55.3} & \textbf{41.3} & \textbf{32.0}   \\
    \bottomrule
  \end{tabular}
  }
  \caption{Ablation of different pre-trained image encoders.}
  \label{tab:modality-encoder}
  \vspace{-1em}
\end{table}

\setlength{\tabcolsep}{3pt}
\renewcommand{\arraystretch}{1.05}

\setlength{\tabcolsep}{4pt}
\renewcommand{\arraystretch}{1.1}
\begin{table*}[t]
\center
\resizebox{0.9\linewidth}{!}{
\begin{tabular}{@{}lcccccccccc@{}}
\toprule
\multicolumn{1}{c}{\multirow{3}{*}{Methods}} & \multicolumn{1}{l}{ \multirow{3}{*}{\#params}} & \begin{tabular}[c]{@{}c@{}}Image\\ Classification\end{tabular} &  \begin{tabular}[c]{@{}c@{}}Object\\ Detection\end{tabular} & \begin{tabular}[c]{@{}c@{}}Instance \\ Segmentation\end{tabular} & \multicolumn{2}{c}{\begin{tabular}[c]{@{}c@{}}Image\\ Captioning\end{tabular}} & \multicolumn{2}{c}{\begin{tabular}[c]{@{}c@{}}Text \\ Retrieval\end{tabular}} & \multicolumn{2}{c}{\begin{tabular}[c]{@{}c@{}}Image\\ Retrieval\end{tabular}} \\ \cmidrule(lr){3-3} \cmidrule(lr){4-4}  \cmidrule(lr){5-5}  \cmidrule(lr){6-7} \cmidrule(lr){8-9} \cmidrule(lr){10-11}
 &  & ImageNet-1k & COCO & COCO & \multicolumn{2}{c}{COCO} & COCO & Flickr30k & COCO & Flickr30k \\ 

 &  & Acc & mAP & mAP & B@4 & CIDEr &  R@1 &   R@1 &  R@1 &  R@1 \\ 
 \midrule

Pix2Seq v2~\cite{pix2seqv2} & 132M  & - & \underline{46.5} & \underline{38.2} & 34.9 & - & - & - & - & - \\
UniTab~\cite{unitab} & 185M  & - & - & - & - & 115.8 &-  & - & - & - \\

Unified-IO\textsubscript{~LARGE}~\cite{unifiedio} & 776M   & 71.8 &  -& - & - &- &-  &-  & - &-  \\
Unified-IO\textsubscript{~XL}~\cite{unifiedio}& 2.9B   & 79.1 & -  & - &  & \underline{122.3}  &- & - & - &  -\\
Flamingo-3B~\cite{alayrac2022flamingo} & 3.2B & -  & - & - & - &-  & 65.9 &  \underline{89.3} & 48.0 & \underline{79.5} \\
Uni-Perceiver\textsubscript{~BASE}~\cite{zhu2022uni_p} & 124M & 79.2 & - & - & 32.0 & - & 64.9  & 82.3 & 50.7 & 71.1 \\
Uni-Perceiver\textsubscript{~LARGE}~\cite{zhu2022uni_p}  & 354M & 82.7 & - & - & 35.3 & - & 67.8  & 83.7  & 54.1 & 74.2  \\
Uni-Perceiver-MoE\textsubscript{~BASE}~\cite{zhu2022uni} & 167M & 80.3 & - & - & 33.2  &-  & 64.6 &  82.1& 51.6  & 72.4 \\
Uni-Perceiver-MoE\textsubscript{~LARGE}~\cite{zhu2022uni}& 505M & \underline{83.4} & - & - & \underline{35.5} & - & \underline{67.9} & 83.6 & \underline{55.3} &  75.9\\
\midrule
Uni-Perceiver-v2\textsubscript{~BASE}  & 308M & 86.3 & 58.6 & 50.6 & 35.4 &  116.9 & 71.8 &  88.1 & 55.6 & 73.8  \\
\multirow{2}{*}{Uni-Perceiver-v2\textsubscript{~LARGE}} & \multirow{2}{*}{446M} & \textbf{87.2} & \textbf{61.9} & \textbf{53.6} & \textbf{36.5} &  \textbf{122.5}  & \textbf{75.0}  & \textbf{89.3} & \textbf{58.5} & \textbf{79.6} \\
& & (+3.8) & (+15.4)  & (+15.4)  & (+1.6)  & (+0.2)  & (+7.1)  & (+0.0)  & (+3.2)  & (+0.1)  \\
\bottomrule 
\end{tabular}}
\caption{Comparison of our Uni-Perceiver  v2  to recent generalist models on six pillar visual and visual-linguistic tasks listed in Tab.~\ref{tab:pillar_tasks}. Note that we only report the results without any task-specific fine-tuning. Uni-Perceiver v2 is the
the first generalist model to support all these pillar tasks and can achieve competitive results without any task-specific adaption.
Some generalist models that only report results with task-specific fine-tuning are not included, \eg, OFA~\cite{ofa} and GIT~\cite{wang2022git}. ``\#params'' is the number of parameters required during model deployment for cross-modal tasks. Results with the best performance are in \textbf{bold}, and previous SoTA results are \underline{underlined}.}
\label{tab:overall}
\vspace{-0.5em}
\end{table*}

\begin{figure*}[ht]
\centering
\includegraphics[width=\textwidth]{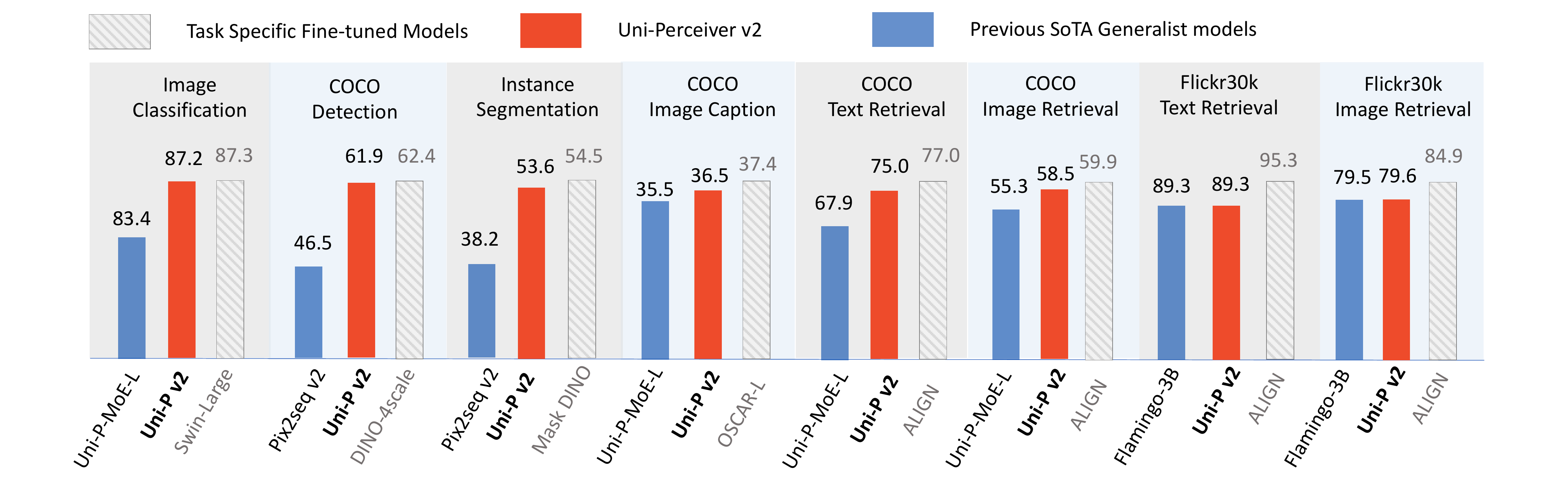}
\vspace{-1.5em}
\caption{Comparison with generalist models and commonly-recognized strong task-specific models on pillar vision and vision-language tasks. 
For generalist models including Uni-Perceiver v2, we only report the results without any task-specific fine-tuning.
Uni-Perceiver v2 (Uni-P v2) is compared with competitive specialized models, \ie Swin-large~\cite{liu2021swin}, DINO~\cite{zhang2022dino}, Mask DINO~\cite{li2022mask}, OSCAR-L~\cite{li2020oscar} and ALIGN~\cite{align}, and previous SoTA generalists, \ie Uni-P-MoE-L~\cite{zhu2022uni}, Pix2seq v2~\cite{pix2seqv2}, and Flamingo-3B~\cite{alayrac2022flamingo}.}
\label{fig:fig2}
\vspace{-1.em}
\end{figure*}

\vspace{0.5em}
\noindent \textbf{Effects of Different Image Encoder Pre-training.~}
By integrating off-the-shelf encoder models, Uni-Perceiver v2 is capable of leveraging existing large-scale pre-trained encoders. To analyze the effects of different pre-training, we employ different pre-trained models for image encoders. For models with supervised pre-training, we employ ResNet-50 pre-trained on ImageNet-1k, on ImageNet-21k, or consecutively pre-trained on ImageNet-1k and COCO. For models with weakly-supervised or unsupervised pre-training, we employ ResNet-50 pre-trained with MoCo v2~\cite{chen2020mocov2} or CLIP~\cite{clip}.
Tab.~\ref{tab:modality-encoder} demonstrates that different pre-training data and methods of image encoders benefit different downstream tasks. Specifically, supervised pre-training methods show the most obvious benefits on downstream tasks similar to it, \eg ImageNet-21k pre-training delivers the best results on ImageNet-1k classification. Besides, the pre-training on large-scale supervised (ImageNet-21k), weakly-supervised or unsupervised data (CLIP and MoCo v2) is more helpful to vision-language tasks such image-text retrieval and image captioning, which possibly thanks to more general representations.

\subsection{Main Results}
To further verify the effectiveness of Uni-Perceiver v2, we incorporate more powerful backbones including Swin-Base and Swin-Large, denoted as Uni-Perceiver-v2\textsubscript{~BASE} and Uni-Perceiver-v2\textsubscript{~LARGE}, respectively. In addition to the tasks included in the ablation studies, we also incorporate instance segmentation on COCO, language modeling on Books\&Wiki, and image captioning / image-text retrieval on YFCC for larger-scale multi-task training.

\vspace{0.5em}\noindent \textbf{Comparison with existing Generalist Models.} 
We list the performance of Uni-Perceiver v2 and other generalist models on pillar vision and vision-language tasks in Tab.~\ref{tab:overall}.
Since generalist models aim to process different tasks with shared architecture and parameters, the task-specific fine-tuning will lose the general modeling ability. We report the performance of the shared models without any task-specific adaptation. 
Specifically, Uni-Perceiver-v2\textsubscript{~BASE} can outperform all previous generalist models on all tasks except the Flickr30k retrieval, even if some methods  have $>10\times$ model parameters, \eg Unified-IO\textsubscript{XL} and Flamingo-3B.
The performance disadvantage on Flicker30k may be due to the use of private data by Flamingo-3B.
Further Scaling up to Swin-Large backbone,  Uni-Perceiver-v2\textsubscript{~LARGE}  obtains the best performance on all tasks.
Thanks to the flexibility of general region proposals,   Uni-Perceiver v2 supports most pillar tasks among generalist models and can achieve competitive results consistently,  which indicates the superior general modeling performance of Uni-Perceiver v2    in both versatility and performance. 

\vspace{0.5em}\noindent \textbf{Comparison with Specialized Models.}
We  compare Uni-Perceiver v2 with  commonly-recognized strong baseline models and previous  SoTA generalist models on the pillar tasks in Tab.~\ref{fig:fig2}. 
The results show that Uni-Perceiver v2 significantly decreases the performance gap between  generalist models  and commonly-recognized strong baselines, which need task-specific fine-tuning.
It can achieve comparable results across all tasks except the retrieval task on Flickr30K, which we suspect is because ALIGN~\cite{align} use 1.8B private image-text pairs, which is much larger than our training data. In contrast, Uni-Perceiver v2 uses only public data for training.

\section{Conclusion}
\label{sec:conclusion}

We propose Uni-Perceiver v2, which is the first generalist model that achieves competitive results on major large-scale vision and vision-language tasks. After being jointly trained on single-modal and multi-modal tasks, Uni-Perceiver v2 achieves competitive performance on a broad range of downstream tasks. As for \textbf{limitations}, our method has not been verified on image generation tasks due to limited computational resources.

% \clearpage
%%%%%%%%% REFERENCES
{\small
\bibliographystyle{ieee_fullname}
\bibliography{egbib}
}

% \clearpage 
\appendix

\section{Architecture Details of the Image Encoder}
\begin{figure}[t]
\centering
\includegraphics[width=.49\textwidth]{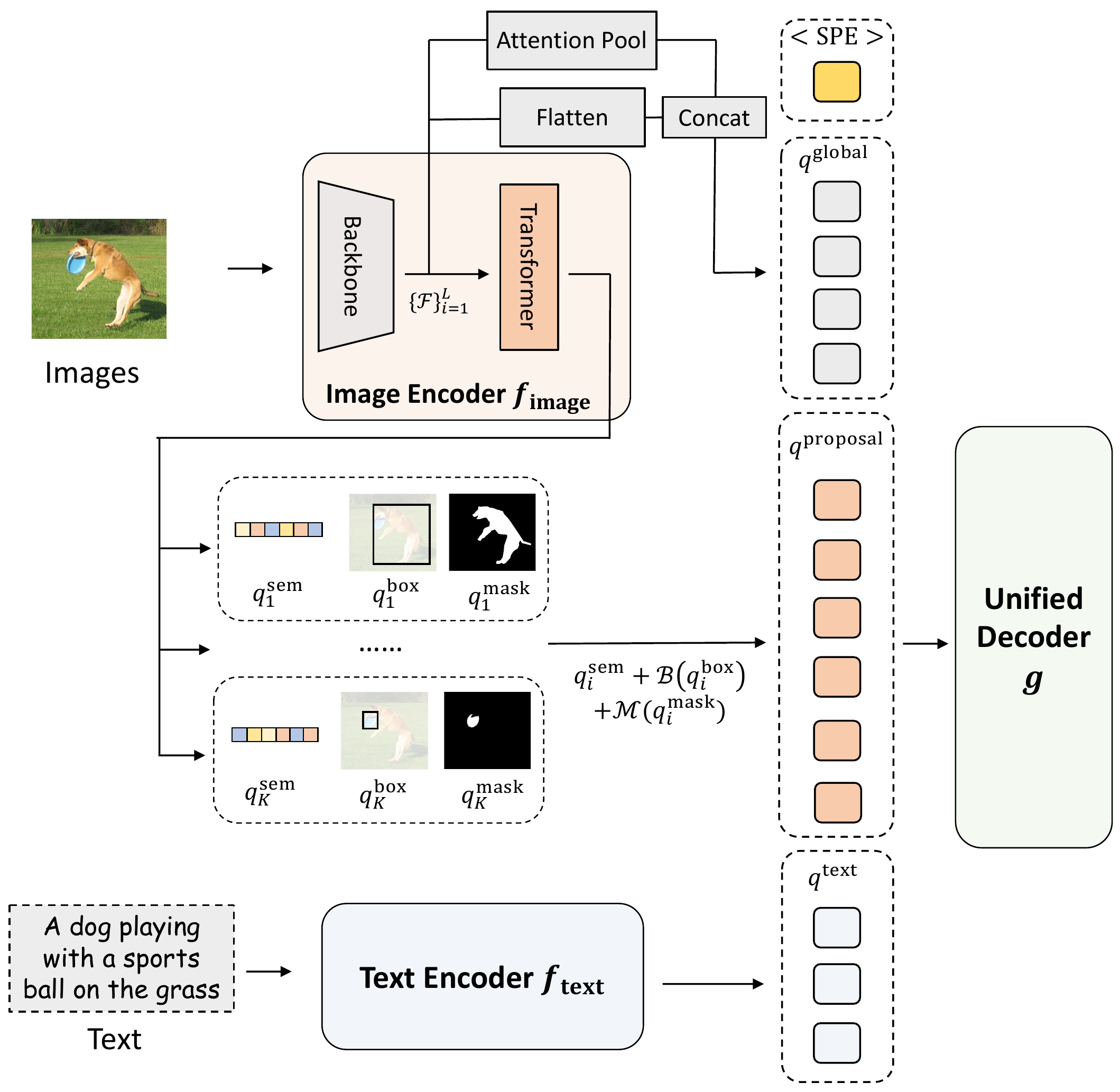}
%\vspace{-0.6em}
\caption{Architecture overview of our Uni-Perceiver v2.}
\label{fig:overallarch}
\vspace{-1.em}
\end{figure}

As shown in Fig. \ref{fig:overallarch}, our Uni-Perceiver v2 consists of three main parts: the image encoder, the text encoder, and the unified decoder. In this section, we describe the architecture details of the image encoder.

 \vspace{0.5em}
\noindent\textbf{Backbone Network.~}Given an input image $x \in \mathbb{R}^{H \times W}$ with height $H$ and width $W$, a backbone network (\eg ResNet~\cite{he2016deep}, Swin-Transformer~\cite{liu2021swin}) is firstly employed to extract the multi-scale feature maps $\{\mathcal{F}_l\}_{l=0}^{L-1}$, where $L=4$ is the number of features scales, and the spatial shapes of the feature maps are $\frac{H}{4} \times \frac{W}{4}$, $\frac{H}{8} \times \frac{W}{8}$, $\frac{H}{16} \times \frac{W}{16}$, and $\frac{H}{32} \times \frac{W}{32}$. The feature maps are transformed by $1 \times 1$ convolutions to match the hidden dimension of the following Transformer-based region proposal network. The transformed feature maps are denoted as $\mathcal{F}'_l$. An additional $3\times3$ stride 2 convolution layer is applied on $\mathcal{F}_3$ to extract a smaller feature map $\mathcal{F}'_4 \in \mathbb{R}^{\frac{H}{64}\times \frac{W}{64}\times d}$. $d=256$ is the hidden dimension of the Transformer.

\vspace{0.5em}\noindent\textbf{Region Proposal Network.~}A Transformer-based region proposal network is applied on top of the multi-scale feature maps to generate regional representations. Specifically, in the 4-scale setting which is adopted by Uni-Perceiver v2, the input of the Transformer encoder is the backbone feature maps except the first scale $\{\mathcal{F}'_l\}_{l=1}^{L=4}$. A deformable Transformer~\citeappendix{zhu2020deformable} encoder is employed to extract multi-scale encoded features $\{\mathcal{F}^{\text{~enc}}_l\}_{l=1}^{L=4}$ whose spatial shapes and dimensions are the same as the corresponding input features. To generate the region proposals, we apply a deformalbe Transformer decoder on the multi-scale encoded features. To construct the $N$ input object queries of the Transformer decoder (\eg $N=900$), we predict the objectness and bounding boxes of each feature pixel in the encoded feature maps $\{\mathcal{F}^{\text{~enc}}_l\}_{l=1}^{L=4}$, and select top-$N$ features based on their objectness. The selected features are added to $N$ randomly initialized object queries as the input of the Transformer decoder, and their locations serve as the initial guess of the bounding boxes of the region proposals. 

The Transformer decoder generates a set of $N$ candidate object proposals $\{ q_j^{\text{sem}}, q_j^{\text{box}}, q_j^{\text{mask}} \}_{j=1}^N$, where $q_j^{\text{sem}} \in \mathbb{R}^d$, $q_j^{\text{box}} \in \mathbb{R}^4$, and $q_j^{\text{mask}} \in \mathbb{R}^{H \times W} $ are the semantic, bounding box, and segmentation mask representations of the $j$-th proposal, respectively. Following Mask2Former~\citeappendix{mask2former} and MaskDINO~\cite{li2022mask}, the segmentation mask representations are obtained by the dot product of the final-layer hidden state of the $j$-th proposal $q_j$ and a per-pixel feature map,
\begin{equation}
    q_i^{\text{mask}} = \text{Upsample}\Big(\text{MLP}(q_i) \odot \mathcal{R}\big(\mathcal{G}(\mathcal{F}_0) + \mathcal{H}(\mathcal{F}_1^{\text{ enc}})\big)\Big),
\end{equation}
where $\mathcal{G}$ is a $1 \times 1$ convolution layer followed by a Group Normalization (GN)~\citeappendix{wu2018group}, $\mathcal{H}$ is a $1 \times 1$ convolution followed by a GN and a bilinear upsampling, and $\mathcal{R}$ is a $3 \times 3$ convolution followed by a GN, a ReLU, and a $1 \times 1$ convolution.

The regional representations are obtained by fusing the semantic, bounding box, and segmentation mask representations,
\begin{equation}
    q_j^{\text{proposal}} = q_j^{\text{sem}} + \mathcal{B}(q_j^{\text{box}}) + \mathcal{M}(q_j^{\text{mask}}),
\end{equation}
where $\mathcal{B}$ denotes the positional encoding of box coordinates.
$\mathcal{M}$ uses adaptive average pooling to scale the mask predictions to the size of $28\times28$. Both $\mathcal{B}$ and $\mathcal{M}$ are followed by linear projections to match the feature dimension. Note that the bounding box and segmentation mask representations are detached before fusing. 

To reduce the computational cost, we predict objectness for each proposal $q_j^{\text{proposal}}$, and select the top-$O$ proposals as the final regional representations. $O$ is set as 200 by default in Uni-Perceiver v2.

\vspace{0.5em}\noindent\textbf{Loss Function.} In non-localization tasks such as image classification, the supervision is applied only on the final predictions of the unified decoder as Eq.~\ref{old loss}, and there is no special supervision for the proposal generation of the image encoder. In localization tasks such as object detection, additional supervisions are applied for the training of the region proposal network. 
Specifically, we adopt the contrastive query denoising in MaskDINO~\cite{li2022mask} for the training of the Transformer decoder. For better convergence of the region proposal network, we predict objectness, bounding box, and segmentation mask for each proposal at the outputs of Transformer encoder and each Transformer decoder layer, and detection losses with binary classification (\ie predicting the objectness instead of classes) are applied to each output as an intermediate supervision.

\section{Implementation Details}

\vspace{0.5em}
\noindent \textbf{Region Proposal Network.}~
The hyper-parameters used in our region proposal network are listed in Tab. \ref{tab:hyperparams}.
These values mainly follow Mask DINO~\cite{li2022mask}, but with small modifications.
The number of candidate object proposals (`num\_queries' in Tab. \ref{tab:hyperparams}) used to generate regional representations is  300 and 900 for the ResNet-50 backbone and Swin backbones, respectively. 
To reduce the computation cost of the unified decoder, the region proposals are 
filtered depending on their objectness scores and only $O=200$ region representations are selected as the input for the unified decoder
(`topk\_queries' in Tab. \ref{tab:hyperparams}). 
Moreover,   
 to save computation cost, the point loss used in  Mask2former~\citeappendix{mask2former} is adopted to calculate mask loss, where the number of sampled points is $112 \times 112$.

\vspace{0.5em}
\noindent \textbf{Unified Decoder.~}As for the Transformer-based unified decoder, a uniform drop rate for stochastic depth is used across all layers and the value is set to 0.1. Unlike Uni-Perceiver series~\cite{zhu2022uni_p,zhu2022uni}, the layer-scale technique~\citeappendix{touvron2021cait} is not enabled since  the instability phenomenon is not observed when the training of the 6-layers unified decoder. In addition, when Conditional MoE is employed in the unified decoder,  the  number of experts in each  layer is set to 8.

\vspace{0.5em}
\noindent \textbf{Data augmentation.~}For all tasks except image detection and segmentation, we apply the data augmentation techniques that are similar to Uni-Perceiver~\cite{zhu2022uni_p}. However, image resolution is set to $384 \times 384$  and $224 \times 224$ for Swin backbones and for ResNet-50 backbone, respectively. And for  object detection and instance segmentation tasks, we first  randomly resize the input image with its shorter side between 200 and 1800 pixels and its longer side at most 2400. Then 
we crop the image to a fixed size of $1600 \times 1600$ during training. For evaluation, the shorter side is set to 1400, and the maximum longer side is set to 1600.

\vspace{0.5em}\noindent\textbf{Others.}~ Tab.~\ref{tab:samplingweight} lists the batch size, sampling weight $s_k$, and scaling factor $\omega_k$ for each task and dataset in the joint training.

\begin{table}[]
\small
    \centering
    \begin{tabular}{l|c}
\hline Item & Value \\
\hline enc\_layers &  6 \\
\hline dec\_layers & 6 \\
\hline dim\_feedforward & \quad\quad2048\quad\quad\quad \\
\hline hidden\_dim & 256 \\
\hline dropout & $0.0$ \\
\hline nheads & 8 \\
\hline num\_queries & 300/900 \\
\hline topk\_queries & 200 \\
\hline enc\_n\_points & 4 \\
\hline dec\_n\_points & 4 \\
\hline cls\_cost\_coef & $2.0$ \\
\hline bbox\_cost\_coef & $5.0$ \\
\hline giou\_cost\_coef & $2.0$ \\
\hline mask\_cost\_coef & $5.0$ \\
\hline dice\_cost\_coef & $5.0$ \\
\hline cls\_loss\_coef & $2.0$ \\
\hline bbox\_loss\_coef & $5.0$ \\
\hline giou\_loss\_coef & $2.0$ \\
\hline mask\_loss\_coef & $5.0$ \\
\hline dice\_loss\_coef & $5.0$ \\
\hline dn\_box\_noise\_scale & $1.0$ \\
\hline dn\_label\_noise\_ratio\quad\quad\quad & $0.5$ \\
\hline
\end{tabular}
    \caption{Hyper-parameters used in our region proposal network.}
    \label{tab:hyperparams}
\end{table}

\begin{table*}[t]
\small
    \centering
\resizebox{0.8\textwidth}{!}{
\begin{tabular}{c|ccccc}

\toprule
task & dataset &\#data & batch size / GPU & sampling weight $s_k$ & scaling factor $\omega_{k}$ \\

\midrule
\multirow{1}{*} Image Classification & ImageNet-1k~\cite{deng2009imagenet} & 1.28M & 28  & 0.1 & 1.0\\
\midrule
\multirow{1}{*} {Object Detection \& Instance Segmentation} & COCO~\cite{lin2014microsoft} & 118K & 1  & 0.25 & 1.0 \\
\midrule
\multirow{1}{*} {Masked Language Modeling} & Books\&Wiki~\cite{zhu2015aligning} & - & 256 & 0.05 & 0.5 \\
\midrule
\multirow{7}{*} {Image Captioning} 
             & YFCC~\cite{yfcc} & 14.8M & 24 & 0.09831 & 0.16385\\
                                        & CC12M~\cite{changpinyo2021cc12m} &  11.1M & 24  & 0.08514 & 0.1419  \\
                                        & CC3M~\cite{sharma2018conceptual} & 3M & 24  & 0.04428 & 0.0738  \\
                                        & Visual Genome~\cite{krishna2017visual} & 108K & 24 & 0.02973 & 0.04955\\
                                        & COCO Caption~\cite{Chen2015MicrosoftCC} & 113K & 24 & 0.0192 & 0.032 \\
                                        & SBU~\cite{ordonez2011im2text} & 830K &24 & 0.02328 & 0.0388  \\
                                        \cmidrule{2-6}
                                        & \textit{sum} & 29.9M & - & 0.3 & 0.5 \\
                                     
\midrule
\multirow{7}{*} {Image-Text Retrieval} 
                                        & YFCC~\cite{yfcc} & 14.8M & 28 & 0.09831 & 0.3277  \\
                                        & CC12M~\cite{changpinyo2021cc12m} & 11.1M & 28 & 0.08514 & 0.2838 \\
                                        & CC3M~\cite{sharma2018conceptual} &3M & 28 & 0.04428  & 0.1476 \\
                                        & Visual Genome~\cite{krishna2017visual} & 108K & 28 & 0.02973 & 0.0991  \\
                                        & COCO Caption~\cite{Chen2015MicrosoftCC} & 113K & 28 & 0.0192 & 0.064 \\
                                        & SBU~\cite{ordonez2011im2text} & 830K & 28 &0.02328 & 0.0776 \\
                                        \cmidrule{2-6}
                                        & \textit{sum} & 29.9M & - & 0.3 & 1.0 \\

\bottomrule
\end{tabular}}
    \caption{
    Tasks and datasets used for our joint training. "\#data" is the amount of visual training samples.   For image captioning and image-text retrieval tasks, a combination of image-text-pair datasets is used for training, which has about 29.9M visual samples after filtering the data overlapping with validation sets.
 To alleviate the data imbalance problem in the combination of image-text-pair datasets during multi-task training,  sampling weight $s_k$ for each dataset is set to be  proportional to the square root of the dataset size, which has demonstrated to be  effective~\cite{zhu2022uni}.} 
    \vspace{0.5em}

    \label{tab:samplingweight}
    \vspace{-1.2em}
\end{table*}

\section{Detection on Novel Categories}
\begin{figure*}[t]
\centering
\includegraphics[width=0.99\textwidth]{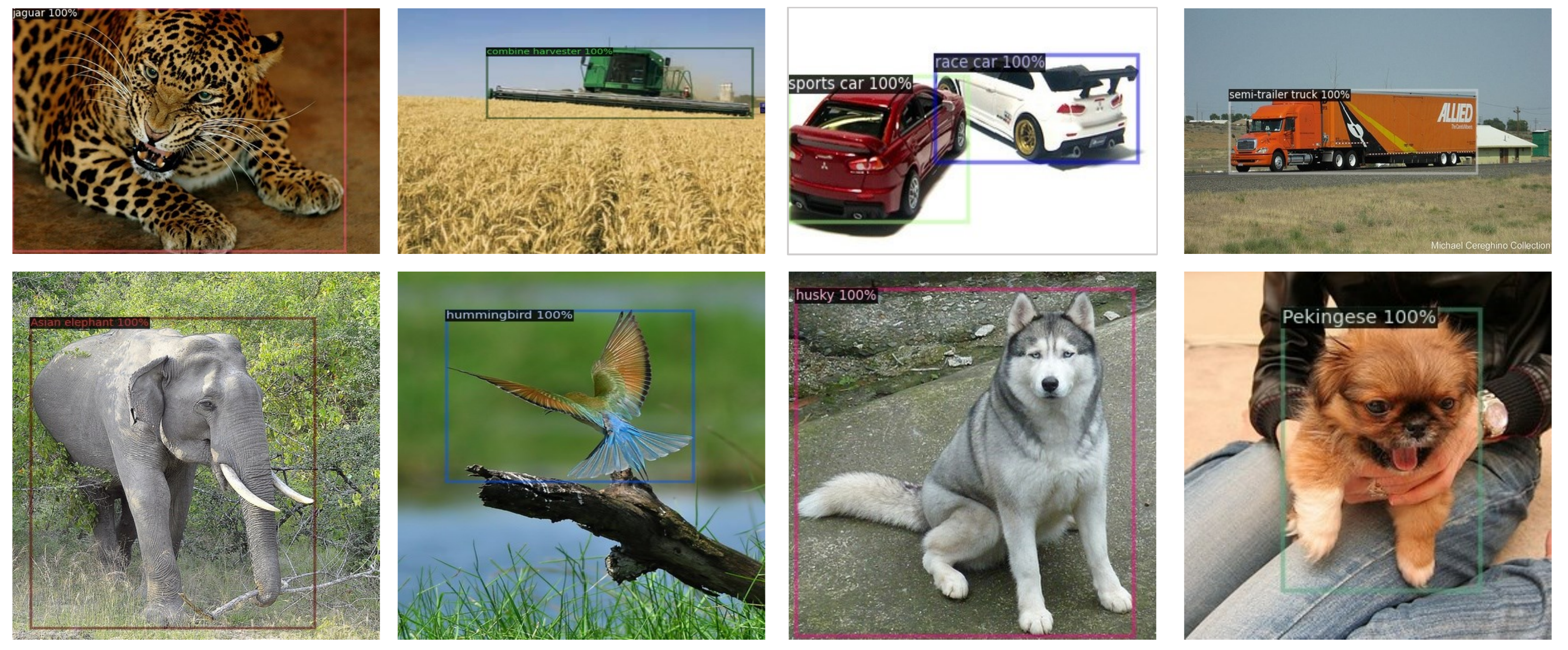}
\caption{Detection results on novel categories. We show the detection results of images from ImageNet-1k validation set. Note that Uni-Perceiver v2 only uses COCO dataset for the training of image detection task, and most classes in ImageNet-1k are not seen in training.}
\label{fig:detvis}
\vspace{-1.em}
\end{figure*}

Thanks to the general task modeling of Uni-Perceiver v2, different tasks can borrow knowledge from each other. For example, object detection task can generalize to novel categories in image classification dataset. Fig. \ref{fig:detvis} shows the detection result of Uni-Perceiver v2 on images in ImageNet-1k validation set whose categories do not exist in COCO dataset. This demonstrates the generalization ability of Uni-Perceiver v2, indicating the benefit of general task modeling.
 
\section{Licenses of Datasets}

\noindent\textbf{ImageNet-1k} \cite{deng2009imagenet} is subject to the ImageNet terms of use \citeappendix{imagenetterms}.

\noindent  \textbf{COCO}~\cite{lin2014microsoft} The images are subject to the Flickr terms of use~\citeappendix{flickr2020terms}.

\noindent\textbf{BookCorpus} \cite{zhu2015aligning} Replicate Toronto BookCorpus is open-source and licensed under GNU GPL, Version 3.

\noindent\textbf{Wikipedia}  Most of Wikipedia's text is co-licensed under the Creative Commons Attribution-ShareAlike 3.0 Unported License (CC BY-SA) and the GNU Free Documentation License (GFDL) (unversioned, with no invariant sections, front-cover texts, or back-cover texts). Some text has been imported only under CC BY-SA and CC BY-SA-compatible license and cannot be reused under GFDL.

\noindent\textbf{YFCC} \cite{yfcc} All the photos and videos provided in YFCC dataset are licensed under one of the Creative Commons copyright licenses.

 \noindent\textbf{CC12M}~\citeappendix{changpinyo2021cc12m} is licensed under  the Terms of Use of Conceptual 12M \citeappendix{cc12mlicense}.
 
 \noindent \textbf{CC3M}~\cite{sharma2018conceptual} is licensed under  the Conceptual Captions  Terms of Use  \citeappendix{cc3mlicense}.
  
 \noindent \textbf{Visual Genome}~\cite{krishna2017visual} is licensed under a Creative Commons Attribution 4.0 International License \citeappendix{vgterms}.
  
\noindent  \textbf{COCO Captions}~\cite{Chen2015MicrosoftCC} The images are subject to the Flickr terms of use~\citeappendix{flickr2020terms}.

 \noindent  \textbf{SBU Caption}~\cite{ordonez2011im2text} The images are subject to the Flickr terms of use~\citeappendix{flickr2020terms}

\vspace{2em}
{\small
\bibliographystyleappendix{ieee_fullname}
\bibliographyappendix{egbib}
}

\end{document}